\documentclass[10pt,twocolumn,letterpaper]{article}

\usepackage{cvpr}              
\usepackage{booktabs}   
\usepackage{multirow}
\usepackage{makecell}
\usepackage{graphicx}   
\usepackage{booktabs}
\usepackage{subcaption}
\usepackage{booktabs}
\usepackage{adjustbox}  
\usepackage{svg}
\usepackage{graphicx}
\usepackage{tikz} 
\usepackage{pifont}
\usepackage{xcolor}
\definecolor{darkgreen}{RGB}{0,153,51}
\usepackage{marvosym}
\usepackage{subcaption}
\usepackage{booktabs}
\usepackage{pifont}
\usepackage[accsupp]{axessibility}

\usepackage[switch]{lineno}  

\definecolor{cvprblue}{rgb}{0.21,0.49,0.74}
\usepackage[pagebackref,breaklinks,colorlinks,allcolors=cvprblue]{hyperref}
\def\paperID{} 
\def\confName{CVPR}
\def\confYear{2026}

\title{Fine-tuning is Not Enough: A Parallel Framework for Collaborative Imitation and Reinforcement Learning in End-to-end Autonomous Driving}
\author{
\setlength{\parskip}{4pt}
Zhexi Lian\textsuperscript{1,†}, Haoran Wang\textsuperscript{1,†}, Xuerun Yan\textsuperscript{1,2,†}, Weimeng Lin\textsuperscript{1}, Xianhong Zhang\textsuperscript{1},\\
Yongyu Chen\textsuperscript{3}, Jia Hu\textsuperscript{1,\Letter}\\
\textsuperscript{1} Tongji University, \textsuperscript{2} Nanyang Technological University, \textsuperscript{3} Chery Automobile \\
{\small
†\ Equal contribution \quad \Letter\ Corresponding author}\\
{\small
The code repository: \url{https://github.com/zhexilian/PaIR-Drive}}
}

\begin{document}
\maketitle
\begin{abstract}
End-to-end autonomous driving is typically built upon imitation learning (IL), yet its performance is constrained by the quality of human demonstrations. To overcome this limitation, recent methods incorporate reinforcement learning (RL) through sequential fine-tuning. However, such a paradigm remains suboptimal: sequential RL fine-tuning can introduce policy drift and often leads to a performance ceiling due to its dependence on the pretrained IL policy. To address these issues, we propose \textbf{PaIR-Drive}, a general \underline{Pa}rallel framework for collaborative \underline{I}mitation and \underline{R}einforcement learning in end-to-end autonomous driving. During training, PaIR-Drive separates IL and RL into two \textbf{parallel} branches with conflict-free training objectives, enabling fully collaborative optimization. This design eliminates the need to retrain RL when applying a new IL policy. During inference, RL leverages the IL policy to further optimize the final plan, allowing performance beyond prior knowledge of IL. Furthermore, we introduce a tree-structured trajectory neural sampler to group relative policy optimization (GRPO) in the RL branch, which enhances exploration capability. Extensive analysis on NAVSIMv1 and v2 benchmark demonstrates that PaIR-Drive achieves \textbf{Competitive} performance of 91.2 PDMS and 87.9 EPDMS, building upon Transfuser and DiffusionDrive IL baselines. PaIR-Drive consistently outperforms existing RL fine-tuning methods, and could even correct human experts' suboptimal behaviors. Qualitative results further confirm that PaIR-Drive can effectively explore and generate high-quality trajectories. 

\end{abstract}
\vspace{-1.2em}
\section{Introduction}
\begin{figure}[t]
  \centering          
  \setlength{\abovecaptionskip}{2pt}   
  \setlength{\belowcaptionskip}{0pt}   
  \includegraphics[width=1.0\columnwidth]{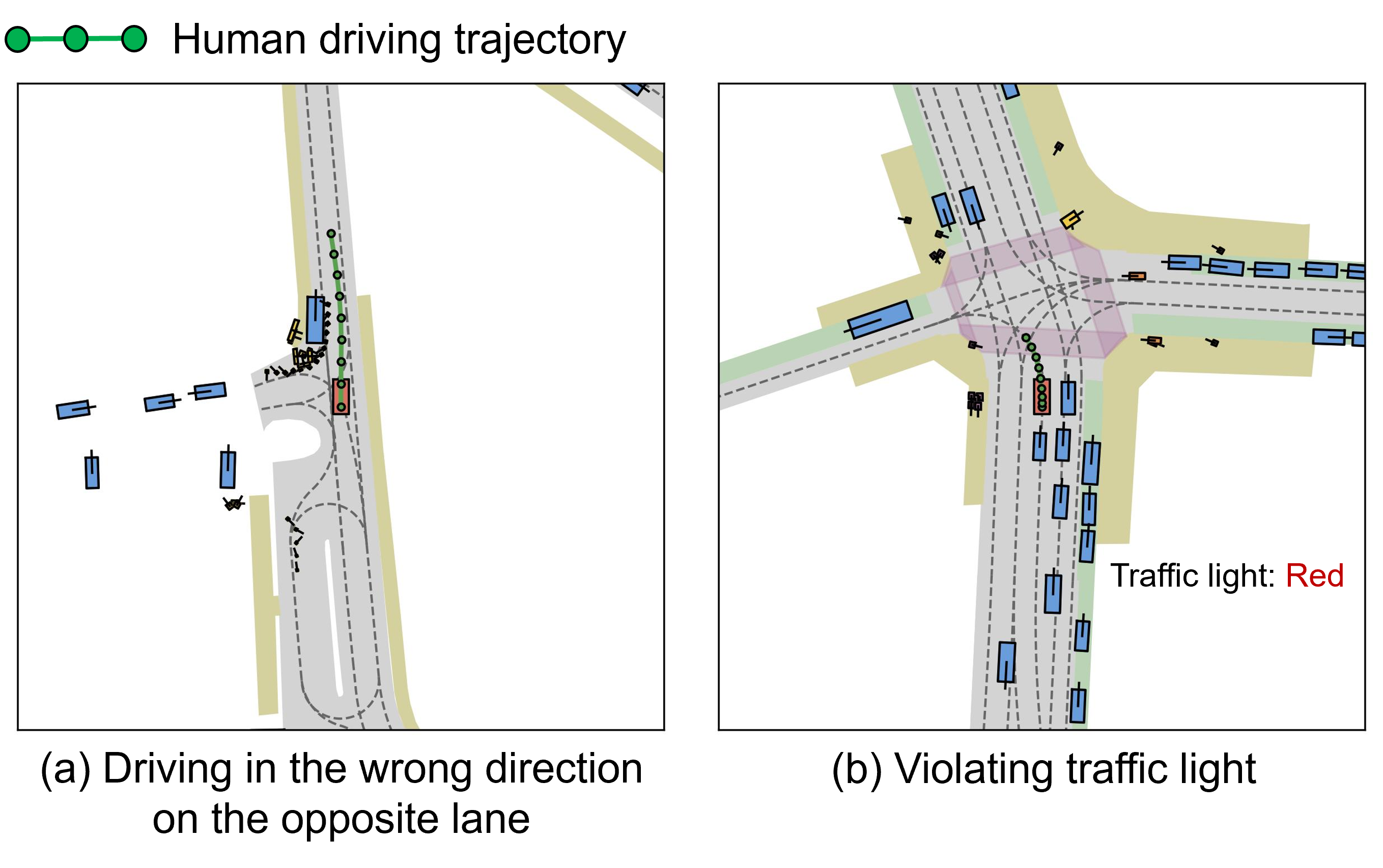}
  \caption{Examples of human's bad behaviors in the real-world dataset NAVSIM. (a) Singapore: The human drives in the wrong direction on the opposite lane; (b) Las Vegas: The human violates traffic light and turns left.}
  \label{fig:intro_humanbad}
\end{figure}

\begin{figure}[t]
  \centering          
  \setlength{\abovecaptionskip}{2pt}   
  \setlength{\belowcaptionskip}{0pt}   
  \includegraphics[width=\columnwidth]{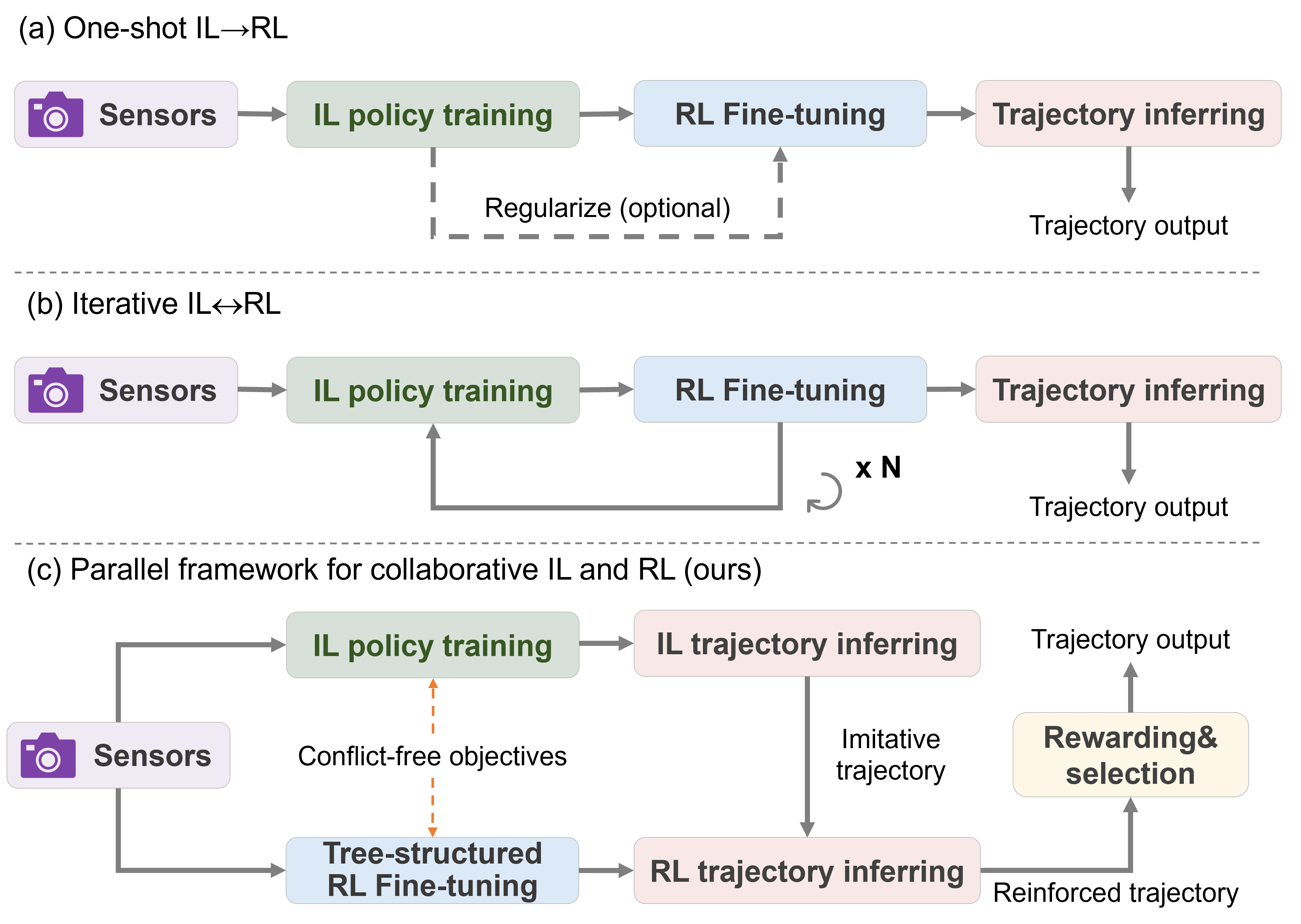}
  \caption{Comparisons of existing training schemes and ours for end-to-end autonomous driving. (a) One-shot IL$\to$RL: IL-based training with subsequent RL fine-tuning; (b) Iterative IL$\leftrightarrow$RL: alternately conducting IL training and RL fine-tuning; (c) Ours parallel framework for collaborative IL and RL.}
  \label{fig:intro_existing_methods}
\end{figure}

{
End-to-end autonomous driving has been developing at a fast pace in recent years. Current mainstream end-to-end autonomous driving methods are typically built upon imitation learning (IL), which aims to mimic human experts’ demonstrations directly from sensor inputs, as exemplified by UniAD \cite{hu2023_uniad}, VAD \cite{jiang2023vad}, DiffusionDrive \cite{diffusiondrive}, etc. Although effective in learning stability, the IL policy is constrained by the quality of human expert demonstrations: (1) IL policy may blindly mimic human's bad behaviors \cite{imitationbootstrappedreinforcementlearning} \cite{imitationNotEnough}. \cref{fig:intro_humanbad} shows some bad behaviors extracted from the large-scale real world dataset NAVSIM, which may misguide the IL policy; (2) IL policy also suffers from low-value driving scenarios. For instance, the dominance of straight-driving scenes (73.9\%) in the nuScenes dataset may cause the IL policy to lack knowledege on dealing with other scenario types \cite{egostatus_2024_CVPR}. Hence, IL-based autonomous driving faces significant challenges in applications.

A potential solution to improve the IL policy is \textbf{fine-tuning} through reinforcement learning (RL). By leveraging reward functions to guide the optimization direction, RL enables the policy to refine its behaviors based on trial-and-error feedback in a closed loop fashion \cite{RLBetter,chen2025researchlearningreasonsearch}. Existing methods have made great progress in sequentially fine-tuning IL policy via RL, which can be categorized as two types: (a) \textbf{one-shot IL$\to$RL}: IL-based training with subsequent RL based fine-tuning \cite{RL1-1-autovla,RL1-2-recogdrive,RL1-3-driveR1,RL1-4-planR1,imitationbootstrappedreinforcementlearning,RL1-5, RL1-6,RL1-7,RL1-8, RL1-17li2025simplevla}, and (b) \textbf{iterative IL$\leftrightarrow$RL}: alternately conducting IL training and RL fine-tuning \cite{RL2-1,RL2-2,RL2-3,RL2-4,RL2-5-IRPO,RL2-6-openvlthinker,RL2-7-liu2023blending,RL2-8-cheng2018fast}. \cref{fig:intro_existing_methods} (a) shows the one-shot IL$\to$RL, a scheme that has gained significant attention after the notable success of DeepSeek-R1 \cite{deepseekai2025}. Some designs include pure on-policy RL fine-tuning through PPO, GRPO, etc, to enhance driving reasoning capability \cite{RL1-1-autovla,RL1-3-driveR1,RL1-4-planR1} or trajectory generation quality \cite{RL1-2-recogdrive,RL1-6}, but these random sampling-based RL fine-tuning results in low sample efficiency when interacting with the environment. Other designs incorporate IL regularization into RL fine-tuning, such as adding an imitation loss to RL rewards \cite{RL1-10,RL1-14-ball2023efficient} or adding expert demonstrations to the replay buffer \cite{imitationbootstrappedreinforcementlearning}. However, this RL fine-tuning scheme continues to face challenges in policy drift (even resulting in lower performance than the original IL policy) and may lead to a performance ceiling due to its dependence on the pretrained IL policy.

Taking into account the above issues, the iterative IL$\leftrightarrow$RL scheme has been proposed, in which IL and RL updates are repeatedly switched, as shown in \cref{fig:intro_existing_methods} (b). This scheme helps to regularize the policy optimization, allowing IL to anchor the policy distribution toward expert behaviors, while RL gradually improves performance through exploration. Common designs include inserting IL updates after several RL updates \cite{RL2-1,RL2-2}, condition-activated RL among IL iterations \cite{RL2-6-openvlthinker,RL2-7-liu2023blending}, adversarial turn-based updates between IL and RL \cite{RL2-9-lee2025unified}. However, IL and RL fine-tuning still conduct upon the same policy network. Given that IL and RL involve different optimization objectives, this fine-tuning scheme could be trapped in a local minimum due to destructive conflicts of inconsistent optimization directions. Hence, a critical scientific question we need to answer is:

\emph{How can we design a unified reinforcement and imitation learning framework to harmonize training objectives, and ultimately surpass IL's prior performance?}

To this end, as shown in \cref{fig:intro_existing_methods} (c), we propose \textbf{PaIR-Drive}, a \underline{Pa}rallel framework for collaborative \underline{I}mitation and \underline{R}einforcement learning in end-to-end autonomous driving. The key insight of PaIR-Drive is breaking the upper performance limit of sequential fine-tuning through our parallel framework. In the parallel IL+RL scheme illustrated in \cref{fig:overall_framework}, the IL branch follows a typical sensor encoding$\rightarrow$perception fusion$\rightarrow$trajectory decoding pipeline. IL's trajectory output is supervised by the human trajectory. Simultaneously, the RL branch takes the BEV feature maps and human expert trajectory as queries to further explore better trajectories. To be specific, we design a tree-structured trajectory neural sampler to predict the trajectory point offsets of driving intentions unseen in human demonstrations. The sampler operates recurrently, where the trajectory tree expansion at each step is conditioned on previous steps. Finally, we use trajectories and their simulated rewards for GRPO to update the policy. As for the inference scheme in \cref{fig:inference}, we can just replace the human trajectory in the RL branch with the trajectory generated by the IL branch, while employing an additional trained reward world model (RWM) to evaluate and select the final plan. The main contributions are listed as follows:

\begin{figure*}[t]
  \centering          
  \setlength{\abovecaptionskip}{1pt}   
  \setlength{\belowcaptionskip}{0pt}   
  \includegraphics[width=2.0\columnwidth]{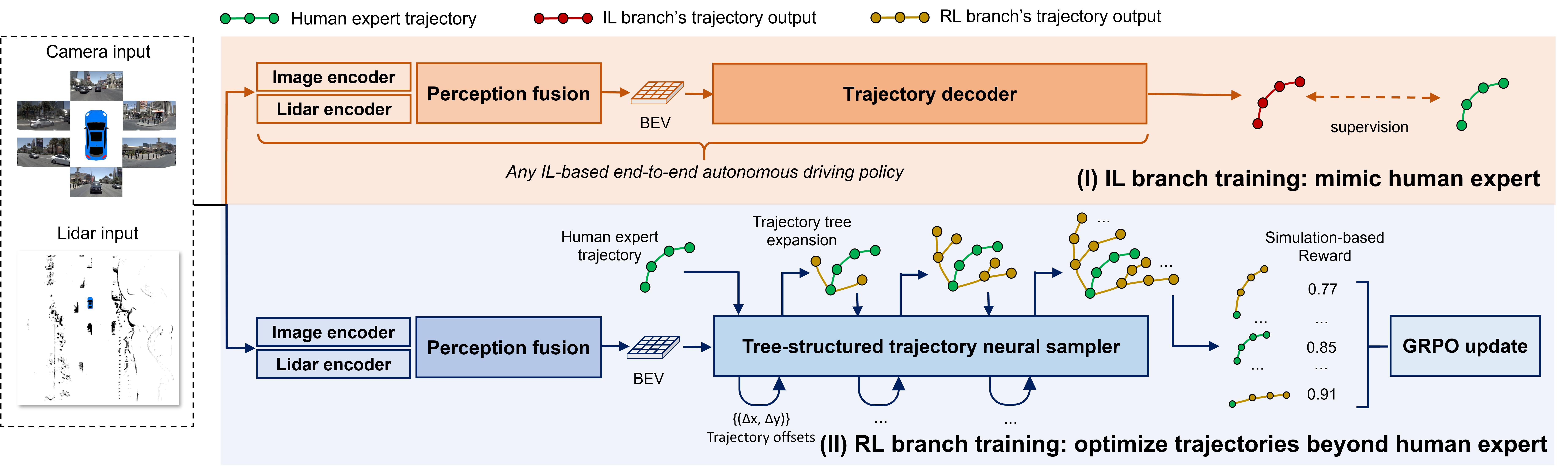}
  \caption{Training process illustration of the parallel scheme of PaIR-Drive. IL branch follows a typical end-to-end planning fashion and is supervised by the human trajectory. Simultaneously, the RL branch builds upon human trajectories and aims to further explore better trajectories. In the RL branch, a tree-structured trajectory neural sampler is designed to recurrently predict the trajectory point offsets of driving intentions unseen in human demonstrations. Finally, we use trajectories and their simulated rewards for GRPO to update the policy.}
  \label{fig:overall_framework}
\end{figure*}

\begin{itemize}
  \item We propose PaIR-Drive, a general parallel framework of IL and RL for end-to-end autonomous driving. By decoupling IL and RL into parallel optimization branches, PaIR-Drive leverages IL to learn human-level driving behavior and RL explores how to surpass human expert performance. Extensive experiments on NAVSIM v1 and v2 benchmarks demonstrate competitive performance of PaIR-Drive.  
  \item PaIR-Drive could serve as a general performance enhancement toolkit. It can be seamlessly integrated into any IL-based autonomous driving method. Its RL branch is built upon human expert demonstrations rather than a specific IL policy, making the framework flexible, adaptive, and widely applicable across different systems.
  \item We introduce a tree-structured trajectory neural sampler to GRPO in the RL branch, which enhances exploration efficiency and improves trajectory quality. Qualitative results showcase the exploration and high-quality trajectory generation capabilities.
\end{itemize}
}
\vspace{-1.2em}
\vspace{4pt}
\section{Related works}
\label{sec:formatting}
{
\setlength{\parskip}{2pt}
\textbf{IL-based end-to-end autonomous driving.} This type of method aims to map sensor inputs to trajectories directly with human expert supervision. UniAD \cite{hu2023_uniad} plays a crucial role in this area as it firstly leverages the advantages of perception and prediction modules for planning. VAD \cite{jiang2023vad, IL-3-chen2024vadv2} introduces vectorized representations for end-to-end planning. Hydra-MDP  \cite{IL-4-li2024hydra} employs knowledge distillation to derive supervision from both rule-based planners and human demonstrations. Diffusion policy \cite{diffusiondrive,IL-5-zhengdiffusion,IL-1-11jiang2025diffvla,IL-1-12fu2025orion,wu2026closed}, world models \cite{IL-6-zheng2025world4drive, IL-7-li2024LAW,IL1-10guan2024world, li2026sgdrive, li2026artdeco, li2025papl}, flow matching \cite{IL-8-xing2025goalflow,IL-9-wang2025flowdrive}, vision-language models \cite{zhang2025cross,weng2026languagegroundeddecoupledactionrepresentation,lin2026harnessingpowerfoundationmodels, song2025hume}, etc, are introduced step-by-step, further enhancing the IL driving policy. However, the performance of IL is constrained by the quality of demonstrations and suffers from low-value driving scenarios.

\noindent\textbf{RL-based end-to-end autonomous driving.} This type of method enables the policy to refine its behaviors based on reward guidance and trial-and-error feedback. Existing methods mainly leverage RL to fine-tune the IL policy, including one-shot IL$\to$RL and iterative IL$\leftrightarrow$RL \cite{li2026plannerrft, liu2026reinforced}. RAD \cite{RL1-6} pretrains a IL policy and fine-tunes it through 3DGS simulation-based RL. Carplanner \cite{RL1-15zhang2025carplanner} combines an auto-regressive RL with an imitation loss to achieve SOTA performance on the challenging large-scale real-world dataset nuPlan. AlphaDrive \cite{RL1-16jiang2025alphadrive} follows a supervised fine-tuning and reinforced fine-tuning scheme and is the first to leverage the advantages of GRPO. PlanRL \cite{RL2-3} switches between IL and RL learning when facing different conditions.  However, given that IL and RL involve different optimization objectives, the fine-tuning scheme could be trapped in a local minimum due to destructive conflicts of optimization directions. Moreover, current IL and RL fine-tuning continues to face challenges in policy drift and may lead to a performance ceiling due to its dependence on the pretrained IL policy.These issues motivate our PaIR-Drive to break the upper performance limit of sequential fine-tuning through our parallel scheme. 
}



\begin{figure*}[t]
  \centering          
  \setlength{\abovecaptionskip}{2pt}   
  \setlength{\belowcaptionskip}{0pt}   
  \includegraphics[width=2\columnwidth]{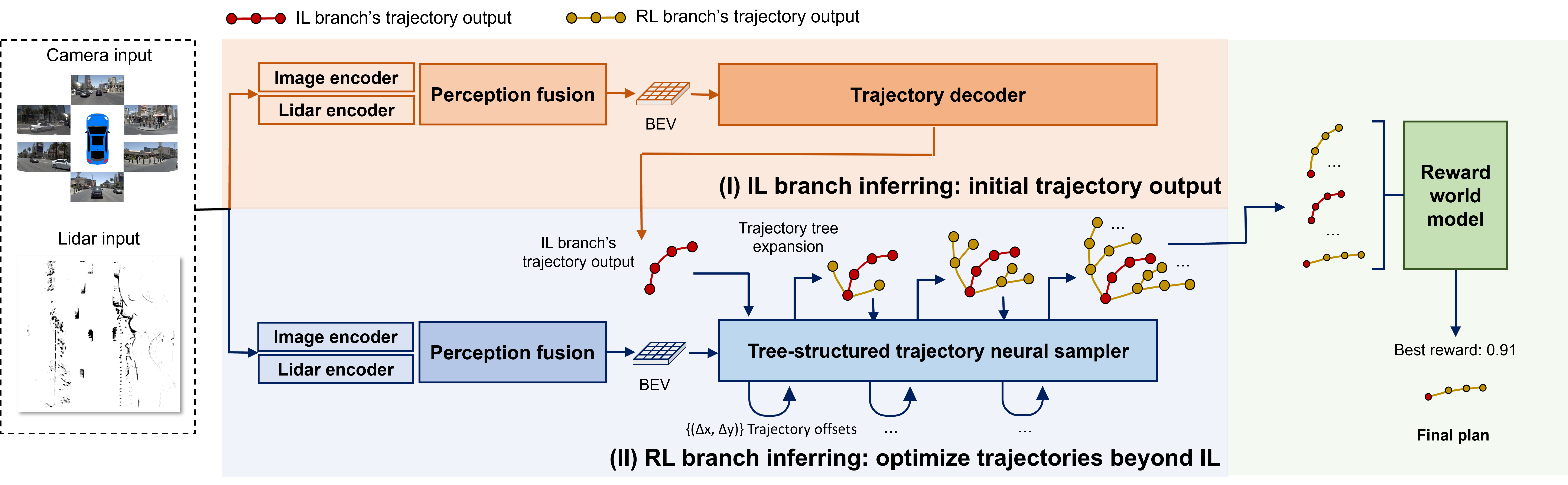}
  \caption{Inferring process illustration of the parallel scheme of PaIR-Drive. Compared with \cref{fig:overall_framework}, we replace the human trajectory in the RL branch with the trajectory generated by the IL branch, while employing an additional trained reward world model to evaluate and select the final plan.}
  \label{fig:inference}
\end{figure*}


\section{PaIR-Drive}
In this section, we introduce four key components of the PaIR-Drive: IL and RL branches formulation (\cref{subsuc:3-1}), the tree-structured trajectory neural sampler (\cref{subsec:3-2}), the training scheme (\cref{sec3-3}), and the inferring scheme (\cref{sec3-4}).
\subsection{Problem formulation}
{
\label{subsuc:3-1}
\textbf{IL branch formulation.} The IL branch of the PaIR-Drive takes ego status, multi-view RGB images from cameras, and point clouds from the lidar as inputs. We use an image encoder and a lidar encoder to obtain perception features $\mathbf{F}_{img}=ImgEncoder(Img),\mathbf{F}_{pcd}=LidarEncoder(Pcd)$. Following a sequential perception fusion and trajectory decoder module, the IL branch generates trajectory output referred to following equations.
\begin{gather}
\tau_{0:T}^{IL}=TajDecoder(\mathbf{F}_{BEV})\\
\mathbf{F}_{BEV}=PercepFusion(\mathbf{F}_{img}, \mathbf{F}_{pcd})\\
\tau_{0:T}^{IL}=\{w_0^{IL},w_{1}^{IL},\dots,w_{T}^{IL}\}
\end{gather}
$\tau_{0:T}^{IL}\in\mathbb{R}^{(T+1)\times3}$ denotes the planned trajectory in the future time horizon $T$, which includes trajectory point $w_t^{IL}$ at each time step $t$. $w_t^{IL}$ contains longitudinal position $x_t$, lateral position $y_t$, and heading $h_t$ in the ego coordinate. The $PercepFusion$ and $TajDecoder$ are borrowed from Transfuser \cite{IL-transfuser}. We would like to emphasize that this IL branch can be replaced by \textbf{any} IL-based autonomous driving policy to generate $\tau_{0:T}^{IL}$.

\noindent\textbf{RL branch formulation.} The RL branch of the PaIR-Drive aims to explore better trajectories. It also takes ego status, multi-view RGB images from cameras, and point clouds from the lidar as inputs and generates BEV feature maps $\mathbf{F}_{BEV}$. Then, we introduce a tree-structured trajectory neural sampler $TreeSampler_i,i\in Intention$. It takes $\mathbf{F}_{BEV}$ and a human expert trajectory as input and predicts the trajectory point offsets relative to the expert trajectory in each intention space (left, right, accelerating, decelerating, etc) in a recurrent manner:
\begin{gather}
\Delta w_{t,i}^{RL}=TreeSampler_i(w_{t,i}^{RL},\mathbf{F}_{BEV})\\
w_{t,i}^{RL}=
\begin{cases}
w_{0}^{Human}, & t=0\text{ if }\ \text{training} \\
w_{0}^{IL}, & t=0\text{ if }\ \text{inferring} \\
w_{t-1,i}^{RL}+\Delta w_{t-1,i}^{RL},  & t > 0
\end{cases} \\
\Delta w_{t,i}^{RL}=\{\Delta x_t,\Delta y_t,\Delta h_t\}\\
\tau_{0:T,i}^{RL}=\{w_{0,i}^{RL},w_{1,i}^{RL},\dots,w_{T,i}^{RL}\}
\end{gather}
$\Delta w_{t,i}^{RL}$ denotes the trajectory point offsets under intention $i$ which include longitudinal position offset $\Delta x_t$, lateral position offset $\Delta y_t$, and heading offset $\Delta h_t$ in the ego coordinate. The RL branch's trajectory output $\tau_{0:T,i}^{RL}$ with different intentions $i$ would expand around the reference trajectory (human expert trajectory in training and IL branch's trajectory output in inferring) with a tree structure. Finally, we use the trajectories and their simulation-based rewards for GRPO to update the policy.
}

\begin{figure}[t]
  \centering          
  \setlength{\abovecaptionskip}{2pt}   
  \setlength{\belowcaptionskip}{0pt}   
  \includegraphics[width=\columnwidth]{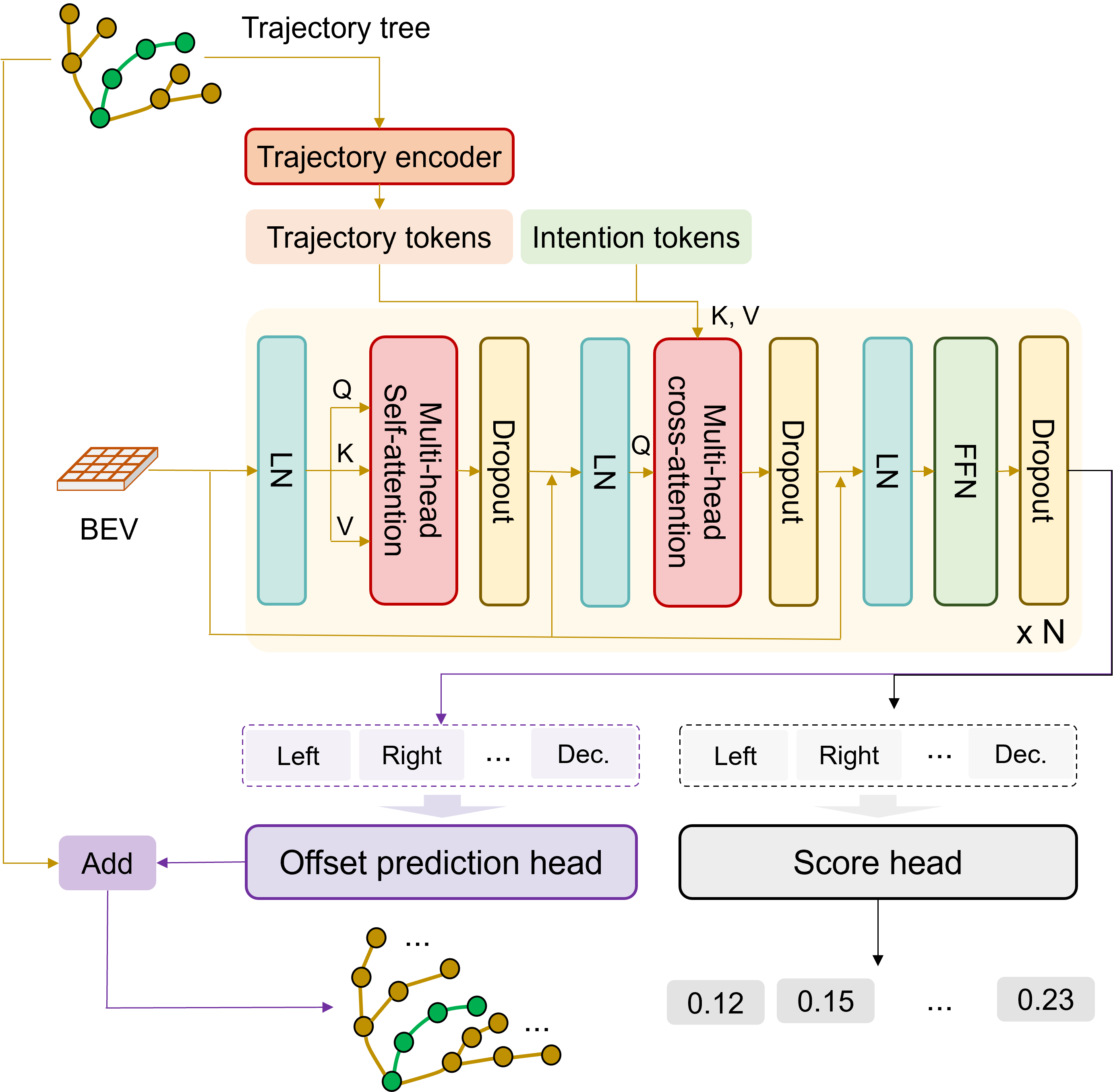}
  \caption{Illustration of the tree-structured trajectory neural sampler with the capability of generating trajectories under different driving intentions unseen in human demonstrations.}
  \label{fig:tree_sampler}
\end{figure}

\subsection{Tree-structured trajectory neural sampler}
{\setlength{\parskip}{1pt}
\label{subsec:3-2}
The core component of the RL branch is the tree-structured trajectory neural sampler. It aims to predict the trajectory point offsets relative to the reference trajectory (human expert trajectory in training and IL branch's trajectory output in inferring) under different driving intentions. The prediction follows a recurrent manner so that trajectories associated with different intentions are progressively expanded, forming a trajectory tree that branches out along the temporal dimension.

\noindent\textbf{Expansion with intentions.} The trajectory offset prediction incorporates $N$ different driving intentions such as Left, Right, Acc.(accelerating), Dec.(decelerating), etc. Assuming that the trajectory tree has already branched out to $M$ trajectories at time step $t$, we need to predict trajectory points offsets with different intentions and expand the original trajectory tree to $M\times N$ at time step $t+1$. Given that the whole trajectory could branch out $T^N$ trajectories which are high-dimensional, we expand trajectories every two steps and candidates with higher exploitative value are selected for further rollouts (More details can be found in the supplementary materials). Hence, the neural sampler could ensure exploration efficiency.

\noindent\textbf{Architecture.} The architecture is illustrated in \cref{fig:tree_sampler}. The input includes $\mathbf{F}_{BEV}$ and the trajectory tree expanded during previous steps. We first generate trajectory tokens $\mathbf{Token}_{traj}\in\mathbb{R}^{(T+1)\times128}$ through a trajectory encoder. Then we concatenate $\mathbf{Token}_{intetion}\in\mathbb{R}^{N\times128}$ produced from intentions with trajectory tokens $\mathbf{Token}_{traj}$. We use a series of multi-head self-attention and cross-attention blocks to capture the latent interaction mechanism between BEV space, trajectory space, and intention space. We predict the trajectory offsets for different intentions using the offset prediction head $OffsetHead$, and predict the log-probability of each offset through the score head $ScoreHead$. In contrast to previous methods, which generate trajectories through randomly sampling \cite{RL1-2-recogdrive,RL1-1-autovla}, the $OffsetHead$ regularizes the offsets prediction range of a specific intention so as to avoid sampling inefficiency (More details can be found in supplementary materials). Finally, we add the offsets prediction to the original trajectory tree and output the trajectory tree at the current step. 
}
\subsection{Training scheme}
{\setlength{\parskip}{1pt}
\label{sec3-3}
This section introduces how we conduct training on the parallel IL+RL framework.

\noindent\textbf{IL branch training.} The IL branch training follows typical supervision fashion as shown in \cref{fig:overall_framework} (I). We optimize the L1 error between the IL branch's trajectory output $\tau_{0:T}^{IL}$ and the human expert trajectory $\tau_{0:T}^{Human}$:
\begin{equation}
    \mathcal{L}_{IL}=\mathrm{L1loss}(\tau_{0:T}^{IL},\tau_{0:T}^{Human})
\end{equation}
The IL branch training is not associated with the RL branch and can be conducted independently.

\noindent\textbf{RL branch training.} The RL branch aims to explore better trajectories. Based on the human expert trajectory, the tree-structured trajectory neural sampler branches out to a trajectory tree with $G$ different intentions $\tau_{0:T}^{RL}=\{\tau_{0:T,i}^{RL},i=1\dots G,i\in Intention\}$. These trajectories are simulated in the NAVSIM simulator online and evaluated by a predefined reward $r_i$, which includes driving safety, drivable area compliance, driving efficiency, driving comfort, etc. The current RL policy $\pi_{\theta}$ is then optimized through GRPO using the normalized group-relative advantage $A_i$.
\begin{gather}
\mathcal{J}_{RL}=\frac{1}{G}\sum_{i=1}^{G}(\mathcal{J}_i-\beta\mathbb{D}_{KL}(\pi_{\theta}|\pi_{old}))\\
\mathcal{J}_i=\mathrm{min}(\frac{\pi_{\theta}(\tau_{0:T,i}^{RL})}{\pi_{old}(\tau_{0:T,i}^{RL})}A_i,\mathrm{clip}(\frac{\pi_{\theta}(\tau_{0:T,i}^{RL})}{\pi_{old}(\tau_{0:T,i}^{RL})},1-\epsilon,1+\epsilon)A_i) \\
A_i = \frac{r_i - \mathrm{mean}\left(\{r_j\}_{j=1}^{G}\right)}{\mathrm{std}\left(\{r_j\}_{j=1}^{G}\right)}
\end{gather}
where $\pi_{old}$ denotes the previous RL policy; $\epsilon$ and $\beta$ are hyperparameters of the clipping range and the KL divergence regularization weight. It can be seen that the RL branch training doesn't depend on the IL branch, so that RL and IL can be conducted in our parallel scheme.
}
\subsection{Inferring scheme}
{
\label{sec3-4}
The inferring scheme is illustrated in \cref{fig:inference}. There are only two differences compared to training the parallel scheme. First, the human expert trajectory is directly replaced by the IL branch's trajectory output. This design is simple but quite beneficial for IL, which has been validated by our experiments. It also means that the RL branch could serve as a general performance enhancement toolkit and be seamlessly integrated into any IL-based autonomous driving method without any retraining. Second, after obtaining the final trajectory tree, we use a reward world model (RWM) to score the trajectories and select the best trajectory as the final plan. The RWM is a lightweight, data-driven alternative to traditional simulators. It predicts the reward $r_i$ and the confidence $conf_i$ of the trajectory $\tau_{0:T,i}^{RL}$ based on current driving commands $c$, and BEV feature maps $\mathbf{F}_{BEV}$:
\begin{equation}
    r_i,conf_i=RWM(\mathbf{F}_{BEV},c,\tau_{0:T,i}^{RL})
\end{equation}
This design can filter those RL branch's exploratory trajectories that are worse than the IL branch's trajectory output.
}

\vspace{-1.2em}
\begin{table*}[t]
\centering
\caption{The human suboptimal behavior improvement capability evaluation on NAVSIMv1 benchmark. The \textcolor{darkgreen}{green score} means the improvement compared with the human driver.}
\small
\begin{adjustbox}{max width=0.75\textwidth} 
\begin{tabular}{*{1}{c}|*{1}{c}|*{5}{c}|*{1}{c}}
\toprule
Data split & Agent &
NC$\uparrow$ & DAC$\uparrow$ & EP$\uparrow$ & TTC$\uparrow$ & C$\uparrow$ & PDMS$\uparrow$ \\
\midrule
\multirow{2}{*}{Human bad v1} &
human & 100.0 & 100.0 & 60.1 & 99.6 & 94.8 & 82.3 \\
& human w/ PaIR-Drive & 100.0 & 100.0 & 62.5 & 99.9 & 97.5 & 83.9(\textcolor{darkgreen}{+1.6}) \\
\midrule
\multirow{2}{*}{Navtest} &
human & 100.0 & 100.0 & 87.4 & 100.0 & 99.6 & 94.7 \\
& human w/ PaIR-Drive & 100.0 & 100.0 & 89.6 & 100.0 & 99.5 & 95.5(\textcolor{darkgreen}{+0.8}) \\
\bottomrule
\end{tabular}
\end{adjustbox}
\label{tab:humanPDMS}
\end{table*}

\begin{table*}[t]
\centering
\caption{The human suboptimal behavior improvement capability evaluation on NAVSIMv2 benchmark. The \textcolor{darkgreen}{green score} means the improvement compared with the human driver.}
\resizebox{0.83\textwidth}{!}{%
\setlength{\tabcolsep}{5pt}
\begin{tabular}{*{1}{c}|*{1}{c}|*{9}{c}|*{1}{c}}
\toprule
Data split & Agent &
NC$\uparrow$ & DAC$\uparrow$ & DDC$\uparrow$ & TLC$\uparrow$ & EP$\uparrow$ &
TTC$\uparrow$ & LK$\uparrow$ & HC$\uparrow$ & EC$\uparrow$ & EPDMS$\uparrow$ \\
\midrule
\multirow{2}{*}{Human bad v2} &
human & 100.0 & 100.0 & 97.0 & 70.4 & 83.2 & 99.6 & 46.2 & 82.7 & 56.6 & 50.0 \\
& human w/ PaIR-Drive & 100.0 & 100.0 & 98.3 & 77.5 & 84.2 & 99.4 & 66.5 & 82.4 & 49.3 & 60.8(\textcolor{darkgreen}{+10.8}) \\
\midrule
\multirow{2}{*}{Navtest} &
human & 100.0 & 100.0 & 99.7 & 97.4 & 87.4 & 100.0 & 87.4 & 98.1 & 90.1 & 90.3 \\
& human w/ PaIR-Drive & 100.0 & 100.0 & 99.9 & 98.0 & 89.6 & 100.0 & 91.7 & 98.1 & 86.4 & 91.9(\textcolor{darkgreen}{+1.6}) \\
\bottomrule
\end{tabular}}
\label{tab:humanEPDMS}
\end{table*}

\section{Experiments}

\subsection{Experimental Setup}
{\setlength{\parskip}{1pt}
\textbf{Benchmarks.} We train and evaluate PaIR-Drive on the large-scale real-world simulation benchmarks NAVSIMv1 \cite{navsim-v1} and NAVSIMv2 \cite{navsim-v2}, which are designed to evaluate autonomous driving performance in complex urban scenarios. The benchmarks include high-quality and challenging scenarios extracted from the Openscene dataset. 

\noindent\textbf{Metrics.} The NAVSIMv1 and NAVSIMv2 benchmarks use PDMS and EPDMS, respectively, to evaluate driving behavior performance. PDMS is a combination of several sub-scores and multiplicative penalties, including No-Collision (NC), Drivable Area Compliance (DAC), Ego Vehicle Progress (EP), Time-to-Collision (TTC), and Comfort (C) \cite{navsim-v1}. 
\begin{equation}
\scalebox{1.05}{$
    \mathrm{PDMS}=\mathrm{NC}\times\mathrm{DAC}\times\frac{5\times\mathrm{EP}+5\times\mathrm{TTC}+2\times\mathrm{C}}{12}$}
\label{eq:pdms}
\end{equation}
EPDMS extends the PDMS, introducing Driving Direction Compliance (DDC), Traffic Light Compliance (TLC), Lane Keeping (LK), History Comfort(HC), and Extended Comfort (EC) \cite{navsim-v2}. $Humanfilter$ can filter those sub-scores that human scores zero, but won't affect the final EPDMS.
\begin{equation}
\scalebox{0.85}{$
\begin{aligned}
\mathrm{EPDMS} 
&= Humanfilter(\mathrm{NC} \times \mathrm{DAC} \times \mathrm{DDC} \times \mathrm{TLC} \\
&\quad \times 
\frac{5\times \mathrm{EP}+5\times \mathrm{TTC}+2\times \mathrm{LK}+2\times \mathrm{HC}+2\times \mathrm{EC}}{16})
\end{aligned}$}
\label{eq:epdms}
\end{equation}

\noindent\textbf{Data splits.} The official navtest split is used for evaluation. Moreover, to evaluate the human experts' suboptimal behaviors improvement capability of the PaIR-Drive, we extract the human bad v1 and human bad v2 splits from the navtest split. The scenarios in human bad v1 are those in which human's PDMS less than 85. The scenarios in human bad v2 are those in which human's EPDMS less than 80.

\noindent\textbf{Implementation Details.} The IL branch and RL branch of the PaIR-Drive both utilize RGB images with a resolution of $1024\times256$ and Lidar's ponit cloud feature with a resolution of $256\times256$. ResNet34 was selected as the perception backbone. The IL branch was trained for 50 epochs using 4 NVIDIA L40 GPUs, each with a batch size of 32. The training used the AdamW optimizer with a learning rate of 1e-4. The RL branch was also trained for 50 epochs using 4 NVIDIA L40 GPUs, each with a batch size of 16. The AdamW optimizer was also adopted, and the learning rate began with 2e-5 and decayed by a cosine schedule. The group size and clip range $\epsilon$ of GRPO were set to 15 and 0.2, respectively. We adopted a dynamic KL divergence regularization weight $\beta$ to stabilize the training process.

\begin{table*}[htbp]
\centering
\caption{Comparison with methods on NAVSIMv1 benchmark. The best and second-best scores are highlighted in \textbf{bold} and \underline{underlined}, respectively. The \textcolor{darkgreen}{green score} means the improvement compared with the origin IL policy. †: using the best-of-N (N=6) strategy following \cite{RL1-1-autovla}.}\label{tab:sotaPDMS}
\small
\begin{adjustbox}{max width=0.85\textwidth} 
\begin{tabular}{*{1}{c}|*{2}{c}|*{5}{c}|*{1}{c}}
\toprule
Method type & Agent & Source &
NC$\uparrow$ & DAC$\uparrow$ & EP$\uparrow$ & TTC$\uparrow$ & C$\uparrow$ & PDMS$\uparrow$ \\
\midrule 
\multirow{8}{*}{IL}
& AutoVLA w/o GRPO \cite{RL1-1-autovla} & NIPS’25 & 96.9 & 92.4 & 75.8 & 88.1 & 99.9 & 80.5 \\
& VADv2 \cite{IL-3-chen2024vadv2} & ICCV’23 & 97.2 & 89.1 & 76.0 & 91.6 & \textbf{100.0} & 80.9 \\
& Transfuser w/o RL \cite{IL-transfuser} & TPAMI’23 & 97.7 & 92.8 & 79.2 & 92.8 & \textbf{100.0} & 84.0 \\
& ReCogDrive w/o RL \cite{RL1-2-recogdrive} & arxiv'25 & 98.3 & 95.1 & 81.1 & 94.3 & \textbf{100.0} & 86.8 \\
& ARTEMIS \cite{baseline-feng2025artemis} & arxiv’25 & 98.3 & 95.1 & 81.4 & 94.3 & \textbf{100.0} & 87.0 \\
& DiffusionDrive \cite{diffusiondrive} & CVPR’25 & 98.2 & 96.2 & 82.2 & 94.7 & \textbf{100.0} & 88.1 \\
& WoTE \cite{IL-2-li2025WOTE} & ICCV’25 & 98.5 & 96.8 & 81.9 & 94.9 & \underline{99.9} & 88.3 \\
& DriveDPO w/o RL \cite{baseline-shang2025drivedpo} & NIPS'25 & 97.9 & 97.3 & 84.0 & 93.6 & \textbf{100.0} & 88.8 \\
\midrule
\multirow{4}{*}{Sequential IL+RL}
& Transfuser w/ GRPO \cite{IL-transfuser} & TPAMI'23 & 98.0 & 94.7 & \textbf{88.5} & 96.6 & \textbf{100.0} & 87.9(\textcolor{darkgreen}{+3.9}) \\
& ReCogDrive w/ GRPO \cite{RL1-2-recogdrive} & arxiv'25 & 98.2 & 97.8 & 83.5 & 95.2 & 99.8 & 89.6(\textcolor{darkgreen}{+2.8}) \\
& DriveDPO w/ DPO \cite{baseline-shang2025drivedpo} & NIPS'25 & 98.5 & 98.1 & 84.3 & 94.8 & \underline{99.9} & 90.0(\textcolor{darkgreen}{+1.2}) \\
& AutoVLA w/ GRPO\textsuperscript{†} \cite{RL1-1-autovla} & NIPS’25 & 99.1 & 97.1 & 87.6 & 97.1 & \textbf{100.0} & 92.1(\textcolor{darkgreen}{+11.6}) \\
\midrule
\multirow{4}{*}{Parallel IL+RL}
& Transfuser w/ PaIR-Drive & ours & 99.1 & 96.1 & 88.1 & 98.2 & 93.1 & 89.7(\textcolor{darkgreen}{+5.7}) \\
& Transfuser w/ PaIR-Drive\textsuperscript{†} & ours & \underline{99.5} & \underline{99.2} & 88.0 & \underline{99.2} & 98.1 & \underline{93.3}(\textcolor{darkgreen}{+9.3}) \\
& DiffusionDrive w/ PaIR-Drive & ours & 99.1 & 97.6 & \underline{88.3} & 98.5 & 94.1 & 91.2(\textcolor{darkgreen}{+3.1}) \\
& DiffusionDrive w/ PaIR-Drive\textsuperscript{†} & ours & \textbf{99.6} & \textbf{99.5} & 88.1 & \textbf{99.5} & 98.6 & \textbf{94.0}(\textcolor{darkgreen}{+5.9})  \\
\bottomrule
\end{tabular}
\end{adjustbox}
\end{table*}


\begin{table*}[htbp]
\centering
\caption{Comparison with SOTA methods on NAVSIMv2 benchmark. The best and second-best scores are highlighted in \textbf{bold} and \underline{underlined}, respectively. The \textcolor{darkgreen}{green score} means the improvement compared with the origin IL policy. †: using the best-of-N (N=6) strategy following \cite{RL1-1-autovla}.}\label{tab:sotaEPDMS}
\resizebox{0.95\textwidth}{!}{%
\setlength{\tabcolsep}{5pt}
\begin{tabular}{*{1}{c}|*{2}{c}|*{9}{c}|*{1}{c}}
\toprule
Method type & Agent & source &
NC$\uparrow$ & DAC$\uparrow$ & DDC$\uparrow$ & TLC$\uparrow$ & EP$\uparrow$ &
TTC$\uparrow$ & LK$\uparrow$ & HC$\uparrow$ & EC$\uparrow$ & EPDMS$\uparrow$ \\
\midrule
\multirow{5}{*}{IL} 
& VADv2  \cite{IL-3-chen2024vadv2} & ICCV’23 & 97.3 & 91.7 & 98.2 & \underline{99.9} & 77.6 & 92.7 & \underline{98.2} & \textbf{100.0} & \underline{97.4} & 76.6 \\
& Transfuser w/o RL \cite{IL-transfuser} & TPAMI’23 & 97.2 & 91.8 & 99.2 & 99.8 & 87.6 & 95.7 & 95.7 & \underline{98.4} & 87.7 & 79.7  \\
& ARTEMIS \cite{baseline-feng2025artemis} & arxiv’25 & 98.3 & 95.1 & 98.6 & 99.8 & 81.5 & 97.4 & 96.5 & \textbf{100.0} & \textbf{98.3} & 83.1 \\
& WOTE \cite{IL-2-li2025WOTE} & ICCV’25 & 98.5 & 96.8 & 98.8 & 99.8 & 86.1 & 97.9 & 95.5 & 98.3 & 82.9 & 84.2 \\
& DiffusionDrive \cite{diffusiondrive} & CVPR’25 & 98.0 & 96.0 & 99.5 & 99.8 & 87.7 & 97.1 & 97.2 & 98.3 & 87.6 & 84.3 \\
\midrule
\multirow{2}{*}{Sequential IL+RL}
& RecogDrive w/ GRPO \cite{RL1-2-recogdrive} & arxiv'25 & 98.3 & 95.2 & 99.5 & 99.8 & 87.1 & 97.5 & 96.6 & 98.3 & 86.5 & 83.6\\
& Transfuser w/ GRPO \cite{IL-transfuser} & TPAMI’23 & 98.0 & 94.7 & 99.3 & 99.8 & \textbf{88.5} & 96.6 & 96.4 & 98.3 & 89.3 & 83.8(\textcolor{darkgreen}{+4.1}) \\
\midrule
\multirow{4}{*}{Parallel IL+RL}
& Transfuser w/ PaIR-Drive & ours & 99.1 & 96.1 & 99.4 & \textbf{100.0} &  88.1 & 98.2 & 96.2 & 94.3 & 74.2 & 86.6(\textcolor{darkgreen}{+6.9}) \\
& Transfuser w/ PaIR-Drive\textsuperscript{†} & ours & \underline{99.5} & \underline{99.0} & \underline{99.6} & \textbf{100.0} & 87.8 & \underline{99.2} & 97.6 & 97.2 & 72.0 & \underline{88.5}(\textcolor{darkgreen}{+8.8}) \\
& DiffusionDrive w/ PaIR-Drive & ours & 99.1 & 97.6 & 99.5 & \textbf{100.0} & \underline{88.3} & 98.5 & 96.9 & 94.8 & 74.0 & 87.9(\textcolor{darkgreen}{+3.6})  \\
& DiffusionDrive w/ PaIR-Drive\textsuperscript{†} & ours & \textbf{99.6} & \textbf{99.5} & \textbf{99.7} & \textbf{100.0} & 88.1 & \textbf{99.5} & \textbf{98.3} & 97.7 & 76.4 & \textbf{89.6}(\textcolor{darkgreen}{+5.3})  \\
\bottomrule
\end{tabular}}
\end{table*}

}
\subsection{Main Results}
{\setlength{\parskip}{2pt}
This section reports the main results of the PaIR-Drive for answering the following questions.

\noindent\textbf{Can PaIR-Drive correct suboptimal human behaviors?} The answer is affirmative. As shown in \cref{tab:humanPDMS}, PaIR-Drive improves upon suboptimal human demonstrations. Using human experts as the base IL policy, PaIR-Drive achieves consistent gains over human performance, with +1.6 and +0.8 PDMS improvements on the Human bad v1 and Navtest splits, respectively. Similarly, as presented in \cref{tab:humanEPDMS}, PaIR-Drive substantially corrects human suboptimal behaviors in metrics such as DDC, TLC, and LK, achieving +10.8 and +1.6 EPDMS gains on the Human bad v2 and Navtest splits, which demonstrates the exploratory capability.

\noindent\textbf{Does PaIR-Drive need retraining when applied to a new IL policy?} The answer is negative. We directly apply PaIR-Drive to different IL policies, including DiffusionDrive and Transfuser, and compare it with competitive pure IL and sequential IL+RL methods. As shown in the PDMS results of \cref{tab:sotaPDMS}, PaIR-Drive improves Transfuser by +5.7 PDMS. PaIR-Drive improves DiffusionDrive by +3.1 PDMS, and achieving a competitive performance of 91.2 PDMS. If we use the best-of-N (N=6) strategy, the performance can reach 94.0 PDMS. Consistent conclusions can be reached in the EPDMS results, as shown in \cref{tab:sotaEPDMS}. PaIR-Drive improves over IL policies, achieving +6.9 and +3.6 EPDMS gains on Transfuser and DiffusionDrive, respectively, and surpassing all other methods across most metrics. Interestingly, the performance enhancement of PaIR-Drive tends to diminish with stronger IL policies, as a well-performed IL policy may leave less room for reinforcement refinement and exploration.

\begin{figure}[b]
  \centering          
  \setlength{\abovecaptionskip}{2pt}   
  \setlength{\belowcaptionskip}{0pt}   
  \includegraphics[width=1\columnwidth]{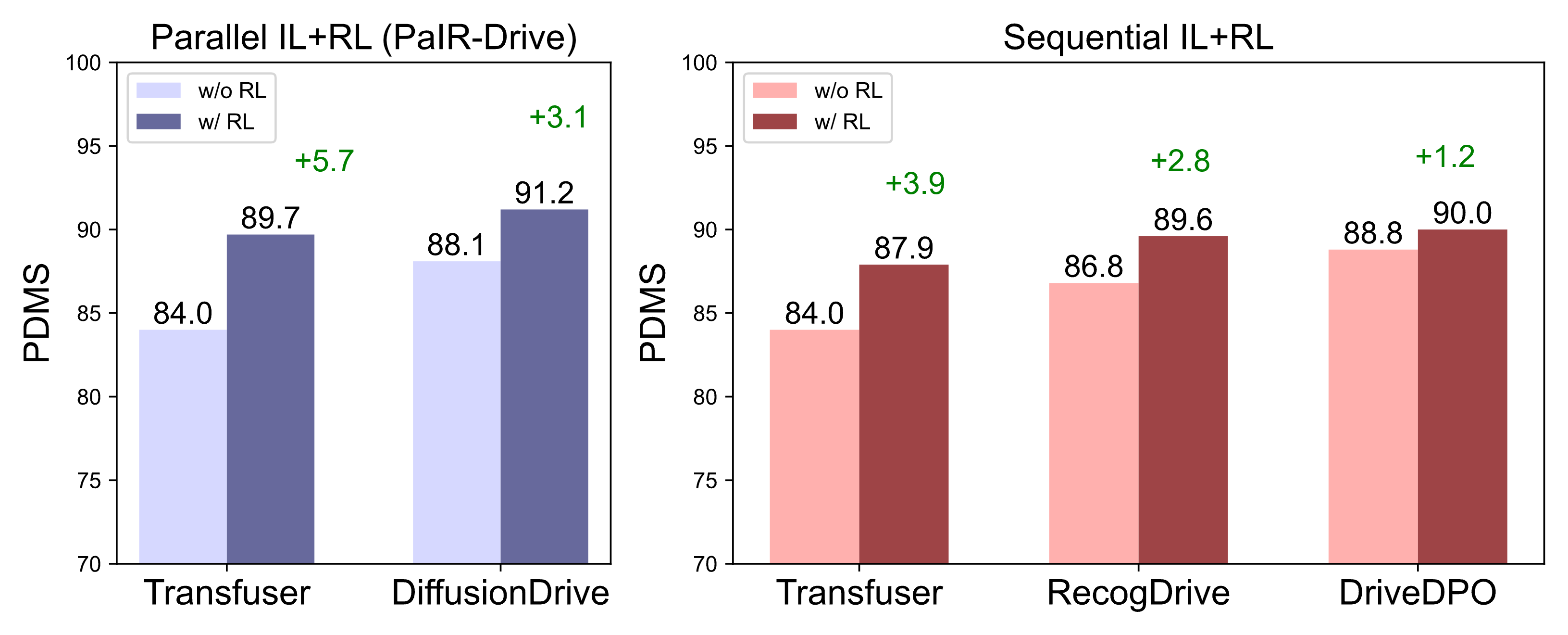}
  \caption{Comparison between our parallel scheme and conventional sequential scheme. Left is the PDMS improvement results of our PaIR-Drive, and right is the results of conventional sequential-RL-based methods.}
  \label{fig:RLtypes}
\end{figure}
}

\subsection{Ablation studies}
{\setlength{\parskip}{2pt}
This section reports the ablation studies on our PaIR-Drive and some interesting findings.

\noindent \textbf{The parallel IL+RL framework matters.} Based on the same Transfuser baseline, PaIR-Drive (\cref{fig:RLtypes} left) reaches 89.7 PDMS, outperforming the sequential counterpart (87.9) in \cref{fig:RLtypes} right. Moreover, with the integration of PaIR-Drive, Transfuser achieves superior performance to stronger IL baselines such as RecogDrive (86.8 PDMS) and DriveDPO (88.8 PDMS), which originally outperform Transfuser under pure IL training. Furthermore, although DiffusionDrive lags behind DriveDPO in the IL-only setting, it surpasses DriveDPO once enhanced by PaIR-Drive, achieving 91.2 PDMS.

\begin{table}[htbp]
\centering
\caption{Ablation studies. The baseline IL policy is Transfuser. offset pred. - predict the trajectory offsets. Traj pred. - generate the whole trajectory directly. All results follow the best-of-N (N=6) strategy \cite{RL1-1-autovla}.}
\label{tab:ablation}

\begin{subtable}[t]{0.5\textwidth}
\centering
\caption{Ablation on tree-structured sampling.}
\label{tab:ablation_on_tree}
\resizebox{0.6\textwidth}{!}{%
\begin{tabular}{c|c|cc}
\toprule
ID & Tree-structured & PDMS$\uparrow$ & EPDMS$\uparrow$ \\
\midrule
1 & \ding{55} offset pred. & 88.8 & 81.6 \\
2 & \ding{55} traj pred. & 87.9 & 83.8 \\
3 & \ding{51} offset pred. & 93.3 & 88.5 \\
\bottomrule
\end{tabular}}
\end{subtable}
\hfill
\begin{subtable}[t]{0.5\textwidth}
\centering
\caption{Ablation on GRPO group number.}
\label{tab:ablation_on_GRPO}
\resizebox{0.6\textwidth}{!}{%
\begin{tabular}{c|c|cc}
\toprule
ID & Group number & PDMS$\uparrow$ & EPDMS$\uparrow$ \\
\midrule
1 & 5 & 89.1 & 80.6 \\
2 & 9 & 89.3 & 81.4 \\
3 & 12 & 93.3 & 86.9 \\
4 & 15 & 93.3 & 88.5 \\
\bottomrule
\end{tabular}}
\end{subtable}
\end{table}

\begin{figure*}[t]
  \centering          
  \setlength{\abovecaptionskip}{2pt}   
  \setlength{\belowcaptionskip}{0pt}   
  \includegraphics[width=1.85\columnwidth]{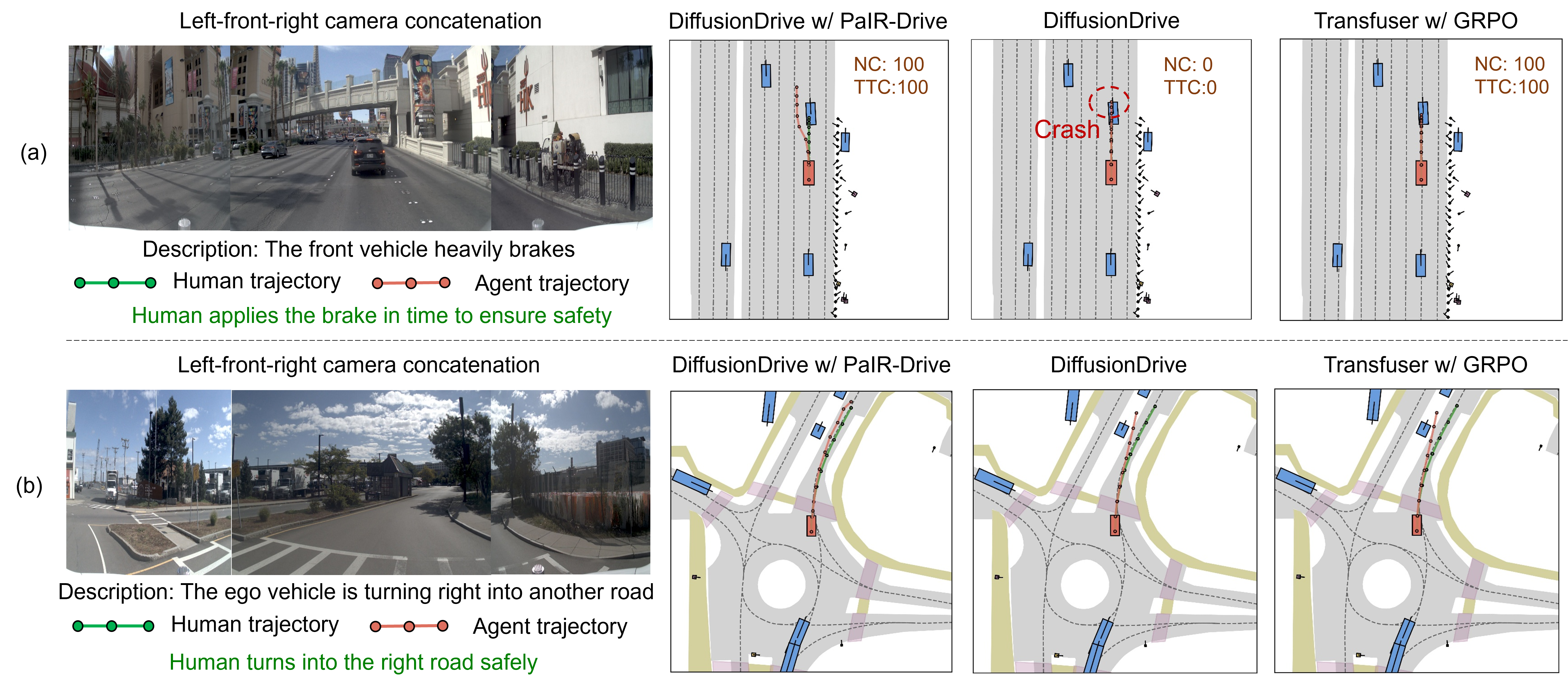}
  \caption{Visualization analysis. We compare PaIR-Drive with DiffusionDrive and Transfuser w/ GRPO. Our PaIR-Drive shows (a) better collision avoidance, (b) more compliant with drivable area.}
  \label{fig:vis}
\end{figure*}

\noindent \textbf{Sampling trajectories with a tree-structured design proves beneficial.} As shown in the last three rows of \cref{tab:ablation} (a), the tree-structured design (ID: 3) yields clear gains over non-structured variants (ID: 1 and 2). It improves PDMS from 88.8/87.9 to 93.3 and EPDMS from 81.6/83.8 to 88.5, highlighting its critical role in stabilizing and enhancing trajectory learning. Interestingly, we found that without the tree structure, trajectory prediction (ID: 2) outperforms offset prediction (ID: 1) in EPDMS but not in PDMS. This is primarily because generating a continuous trajectory ensures driving comfort, which is captured in EPDMS.

\noindent \textbf{More GRPO group number of trajectories helps.} \cref{tab:ablation} (b) shows that increasing the GRPO group number leads to consistent performance gains. As the group number grows from 5 to 15, PDMS improves from 89.1 to 93.3, and EPDMS rises from 80.6 to 88.5. This demonstrates that due to our tree-structure design, larger group size leads to richer trajectory diversity, enabling more effective policy optimization under the parallel framework.

\noindent \textbf{RWM is not the decisive factor.} As shown in \cref{tab:ablation_RWM}, directly applying IL + RWM yields only limited gains (+2.7 EPDMS), whereas our PaIR-Drive + RWM achieves substantially better performance (+5.3 EPDMS). It demonstrates that the improvement does not come from RWM re-ranking alone.

\begin{table}[t]
\centering
\caption{Ablation on the dependance of RWM. The pretrained IL policy is Diffusiondrive.}
\small
\begin{adjustbox}{max width=0.7\textwidth} 
\begin{tabular}{*{1}{c}|*{1}{c}|*{2}{c}}
\toprule
ID & Agent & PDMS$\uparrow$ & EPDMS$\uparrow$ \\
\midrule
1 & Vanilla IL & 88.1 & 84.3\\
2 & IL + RWM & 90.2 & 87.0\\
3 & PaIR-Drive + RWM & 94.0 & 89.6\\
\bottomrule
\end{tabular}
\end{adjustbox}
\label{tab:ablation_RWM}
\end{table}

\subsection{Visualization analysis}
{\setlength{\parskip}{2pt}
This section reports three representative cases to validate the advantages of our PaIR-Drive.

\noindent \textbf{Case (a): Better avoiding collision.} As shown in \cref{fig:vis} (a), the front vehicle brakes heavily. The human expert and the Transfuser w/ GRPO both brake in time to ensure safety. However, the DiffusionDrive fails to decelerate and collides with the front vehicle. Interestingly, our PaIR-Drive learns to change lane proactively, balancing collision avoidance with improved driving mobility. The result confirms the advantage of the multi-intentions trajectory expansion of our PaIR-Drive.

\noindent \textbf{Case (b): More compliant with drivable area.} As shown in \cref{fig:vis} (b), the human driver successfully pass through the roundabout. DiffusionDrive and Transfuser w/GRPO show unstable trajectories, drifting slightly during the entry phase. In contrast, PaIR-Drive follows a clean and well-aligned arc into the correct lane, maintaining stability and consistency throughout the maneuver. This case confirms the advantage of our multi-intention trajectory expansion, enabling PaIR-Drive to better reason over complex
geometric structures like roundabouts.

\vspace{4pt}
\section{Conclusion}
{
In this paper, we introduce \textbf{PaIR-Drive}, a general \underline{Pa}rallel framework for collaborative \underline{I}mitation and \underline{R}einforcement learning in end-to-end autonomous driving. By decoupling IL and RL into parallel optimization branches, PaIR-Drive leverages IL to learn human-level driving behavior and RL to explore how to surpass human expert performance. PaIR-Drive could serve as a general performance enhancement toolkit and be seamlessly integrated into any IL-based autonomous driving method without any retraining. A tree-structured trajectory neural sampler is introduced to GRPO in the RL branch, which further enhances exploration efficiency and improves trajectory quality. Our PaIR-Drive achieves competitive performance on NAVSIM v1 and v2 benchmarks. Visualization analysis showcases the exploration and high-quality trajectory generation capabilities of PaIR-Drive. Overall, our key insight is breaking the upper performance limit of sequential fine-tuning through our innovative parallel framework. We hope our work could inspire further research in different IL+RL schemes of end-to-end autonomous driving. 
}
\section{Acknowledgment}
{This paper is partially supported by National Natural Science Foundation of China (Grant No. 52372317 and 52302412), Yangtze River Delta Science and Technology Innovation Joint Force (No. YDZX20233100004027), Shanghai Automotive Industry Science and Technology Development Foundation (No. 2404), Xiaomi Young Talents Program, the Fundamental Research Funds for the Central Universities (22120230311), and Tongji Zhongte Chair Professor Foundation (No. 000000375-2018082).}

{
\small
\bibliographystyle{ieeenat_fullname}
\bibliography{main}}


\end{document}